# Decision Making Using Rough Set based Spanning Sets for a Decision System

Nidhika Yadav[a]

**Abstract:** Rough Set based concepts of Span and Spanning Sets were recently proposed to deal with uncertainties in data. Here, this paper, presents novel concepts for generic decision-making process using Rough Set based span for a decision table. Majority of problems in Artificial Intelligence deal with decision making. This paper provides real life applications of proposed Rough Set based span for decision tables. Here, novel concept of span for a decision table is proposed, illustrated with real life example of flood relief and rescue team assignment. Its uses, applications and properties are explored. The key contribution of paper is primarily to study decision making using Rough Set based Span for a decision tables, as against an information system in prior works. Here, the main contribution is that decision classes are automatically learned by the technique of Rough Set based span, for a particular problem, hence automating the decision-making process. These decision-making tools based on span can guide an expert in taking decisions in tough and time-bound situations.

## 1. Introduction

Uncertainty [3] in data and decision making have always posed challenges for mankind to take right decisions. There have been various studies in decision making in uncertainty [3, 5, 16], still much more progress and research is as much needed as before. Not only to assist humans but also to help machines take their own decisions, to some extent. Fuzzy Sets [3] and Rough Sets [16] are two popular techniques to deal with uncertainty occurring in data and hence in decision making.

The concept of uncertainty handling based on Fuzzy Sets [3] rely on membership value of each element of a universe to a set X. While the uncertainty handling via Rough Sets deal with subsets of universe and the granulation of universe in equivalence classes. The partial and complete belonging of these information granules in a subset determines the upper and lower approximations of the subset. The boundary region consists of those granules which are not in lower approximation and have some information common with the set under consideration.

Rough Sets have been used since its inception in 1982 by Pawlak [16, 17] in variety of applications. Some of the key areas where Rough Sets have been applied are feature selection [2, 11, 12, 19], classification [13], text summarization [20, 21], financial data analysis [18], medicine [9], data mining [8, 14], clustering [15], information retrieval [4] to mention a few.

The motivation of this paper is three-fold. Firstly, it presents real-life problems that require AI based automation using previously defined concept of Rough Set based span. Secondly, to propose the concept of span for a decision table, previous work, deals only with an information system with X, a subset of U, as a variable. While here the full decision class is a variable. The aim is to direct the search of right decision class via the concept of finding best span for the decision system. This can be considered as a stochastic search technique automating the decision category of each object of the universe. And hence assist in automatic unsupervised directed learning.

[a] email address: nidhikayadav678@outlook.com

The paper is organized as follows. Section 2 discusses previous work pertaining to the proposed decision-making system and introduces the problem of flood rescue and relief team dispatch. The proposed novel concepts of span of a decision table are introduced and illustrated for example and applications in Section 3. Section 4 presents the effect of feature selection via reducts on the span of a decision table. Conclusion and future work are given in Section 5.

## 2. Previous Work

Rough Set or any decision-making system depends on an information system, in case of unsupervised learning and on a decision system, for a supervised learning system. The information system consists of measurements or values of objects of universe for various features called attributes. The information system can be represented in a tabular form in which the rows depict the objects and the columns represents the properties, measurements, physical attribute or any other representational data about the objects. All objects are typically represented with the same attributes, otherwise a null symbolic value is inserted in the place, which is dealt in machine learning, AI and data mining in altogether separate way.

Yadav, N. et al. (2019) defined span and spanning set for a universe U and a set of attributes P for an information system. Here, the paper provides a modified version of the same definition explicitly for a complete information system, complete in sense that full attribute set R is considered in the definition. Hence in this definition the only variable is X, since attributes are fixed as the complete attribute set. Span of a set X given an information system is defined as follows.

**Definition 1**. Given a universe $U$ and an Information System $(U, R)$. The span of a subset $X$ of $U$ is defined as:
$\delta_X = \left(w_1 * \frac{|RX|}{|U|} + w_2 * \frac{|BN_R(X)|}{|U|}\right)$, where $w_1, w_2 \in [0,1]$, $w_1+w_2=1$.

This definition is needed as in many applications one does not require a subset of attributes, while in other problems attribute subsets are required to reduce the computational times and varying requirements. Hence, both definitions have their own needs and advantageous. This definition has lot of advantageous especially in unsupervised learning. Spanning set is a subset of universe which maximizes the span, given the information table. Further, the spanning set may not be unique for a problem. Some novel applications on real world decision problems of the concept can be illustrated as follows:

1. Finding a representative group for a task such as selection of leaders of a labor union for elections. This problem can be viewed as finding a group which represent the entire work force of an organization. This is an unsupervised learning problem and can be represented in form of an information table, wherein, each row represents the employee and each column represent the attributes describing the broad categorization of the employees. The aim is to select a group of leaders which can cater for most demands of employees and at the same time are uniquely distinct and minimal in number. This problem is a decision-making problem can be considered as the problem of finding spanning set of this information.

2. Selecting a team to represent a group in a company. The group can be a software division and a team of size n is to be made to represent this software division. The group needs to represent the variety and vividity of the complete division. This problem can be considered as a subset selection problem, which can be solved by computing a spanning set.

3. Relief Dispatch. This is a relief dispatch problem which can be a flood relief dispatch or any other natural calamity relief dispatch problem. This problem can be viewed as the problem in which relief in form of food supplies, medicines, emergency aid can reach to as many places as possible. Consider a flood affected region K, this region can have various places or people that are stranded alone in floods, let us call them K1, K2,….,Kn. These places K1, K2,….,Kn can have other people, cattle's and children struck in floods. They may require food, boats, choppers, medicines to mention a few. Given the situation the amount of food, rescue teams are limited in number. So, how does on decides how to send rescue teams and to which locations? How much food is to be distributed and to whom? This problem can be considered as finding the spanning set to form the rescue team having a limited number say, m. These m teams have to reach the n places struck in a flood. The n places can be small regions to even complete village.

The aim is to save as many lives as possible and to maximum possible. There are some regions which are to be surely covered in this task and some regions that may possibly be covered. Assigning these m rescue teams to n places, where some places have to be surely covered due to some pre-determined categories which for rest have to be covered so as to save as many people as possible. The task is to find a subset X of {K1, K2…,Kn} for the first visit of the rescue team which maximizes the span, hence covers must covered areas. Consequently, form the optional list for those areas that has maximum coverage of life and resources after the first visit. Say {K11,…K1n} are selected in this process as the spanning set. In the subsequent visits the {K1, K2…,Kn} - {K11,…K1n} is taken as the new universe and a new constraint as new rescue teams and new distribution of food task comes in picture.

There are problems in which information system is not required here. A partition of universe is enough to define the problem. This partition of universe can be attained by expert knowledge or computations. As an example, consider the flood relief team dispatch problem. The experts can club in the regions {K1, …, Kn} into equivalence classes, so that regions with same entry point for rescue are together. The certainly covered regions have to be included in spanning set by partitioning suitably while others can be clubbed by the expert handling flood relief program. The spanning set of size m is determined to meet the requirements and constraints. The problem can be solved as an optimization problem as suggested in Yadav et.al. (2019).

This motivates us to define a span and spanning set for a partition of a universe of objects. The proposed definition is given as follows:

**Definition 2**. Given a partition of universe $U$ into disjoint equivalence classes $R$. The span of a subset $X$ of $U$ is defined as:
$$\delta_X = \left(w_1 * \frac{|\underline{R}X|}{|U|} + w_2 * \frac{|BN_R(X)|}{|U|}\right), \text{ where } w_1, w_2 \in [0,1], w_1 + w_2 = 1.$$

The key difference between definition 1 and definition 2 is that in definition 2 explicit attribute set is not required, only knowledge of universe is essential in this regard. Example of use of this definition lies in many expert systems, where the expert is able to partition the universe in equivalence classes, however, the attributes cannot be assigned as many entries may contain a null value or is unmeasurable. The above example of decision making in *flood relief dispatch* is an ideal example, where the experts can partition the universe of locations that require relief and need to select a subset of these locations, through which relief can reach maximum people and to essential places as well. All this keeping in mind the constraints, viz. the number of relief teams, and hence the number of locations that can be reached in one go.

In the Section 3, the span of a decision table (system) is proposed and its properties and applications are elaborated in detail.

## 3. Span of a Decision System

Span of an information system has been well studied and discussed, here we lay emphasis on span of a decision system tabulated as a decision table. A decision system is a knowledge base system, that apart from containing, the knowledge about the objects of universe in form of attributes, also have a decision class for each object. The decision class is the category to which an object belongs. It can be a binary classification problem or a multi-class problem. Such systems are present in problem domains in the area of supervised learning. One may learn from such system rules or they can be fed into a machine learning algorithm.

Now, here we define a spanning capability of a subset X of U given the attributes R describing it and attributes D forming the decision category of the objects. Given a decision system $(U, R, D)$ the span of this system is proposed to be defined as follows.

**Definition 3.** Given a decision system $(U, R, D)$, $P \subseteq R$, $U/D = \{D1, D2, \ldots, Dr\}$. The span of D is defined as the weighted average,

$$\Delta_{P,D} = \frac{1}{|U/D|} \sum_{X \in U/D} \left( w1 \frac{|\underline{P}X|}{|U|} + w2 \frac{|BN_R(X)|}{|U|} \right), w1, w2 \in [0,1], w1 + w2 = 1.$$

The span of a decision system defines the ability of the decision attributes to cover the knowledge represented by the knowledge base of the decision system.

**Definition 4.** The complete span of a decision system $(U, R, D)$, $P \subseteq R$, based on the given attribute set is given by:

$$\Delta_{P,D}^{Complete} = \frac{1}{|P|} \sum_{a_i \in P} \Delta_{a_i, D}$$

This is illustrated with the following example.

**Example 1.** Consider the Decision System as given in Table 1, for a decision making in flood relief operation. Here $U = \{o1, o2, o3, o4, o5, o6\}$, $R = \{a1, a2, a3, a4, a5\}$. Where *a1* refers to location with values A and W for approachability, *a2* refers to health of Good(G), Bad(B) and A(average), *a3* attribute is for food supplies in regions with values Y(yes) and No(N), *a4* for water supplies with values No(N) and Yes(Y) and the last attribute *a5* for availability of utility in regions with values No(N) and Yes(Y). The decision values is marked by the expert who have to sign off a rescue mission in the flood effected regions *{o1, o2, o3, o4, o5, o6}*. The decision values taken in this example are Team1(T1) and Team2(T2), depicting the experts decision as being T1 send on location or T2 send for the rescue team to reach in first phase.

Let *w1*=0.3 and *w2*= 0.7 be fixed for the problem. These weights can be leaned using various techniques. Yadav et al. (2019) used PSO algorithm [6, 19] to learn these weights. These values of weights depict that boundary region is given a preference while keep surely covered elements in lower approximation at par.

|     | a1=loc | a2=health | a3=food | a4=water | a5=utility | D  | D1 |
|-----|--------|-----------|---------|----------|------------|----|----|
| o1  | A      | G         | Y       | N        | N          | T1 | T1 |
| o2  | A      | B         | N       | N        | N          | T2 | T1 |
| o3  | W      | G         | N       | N        | N          | T1 | T2 |
| o4  | W      | A         | N       | Y        | Y          | T1 | T2 |
| o5  | A      | A         | N       | Y        | N          | T2 | T2 |
| o6  | A      | A         | Y       | Y        | N          | T2 | T2 |

Table 1. Decision Table for Example 1

Let us consider *U/D1 = {{o1, o2}=Urgent, {o3, o4, o5, o6}=Not-Urgent}*

The following are the indiscernibility relation w.r.t single attributes.

*U/a1 = {{o1,o2,o5,o6}, {o3, o4} }*
*U/a2 = {{o1,o3},{o2},{o4,o5,o6}}*
*U/a3 = {{o1, o6}, {o2, o3, o4, o5}}*
*U/a4 = {{o1, o2, o3}, {o4, o5, o6}}*
*U/a5 = { {o1, o2, o3, o5,o6}, {o4}}*

The complete span of the decision system is given by $\Delta_{R,D}^{Complete} = \frac{1}{|R|}\sum_{a_i \in R} \Delta_{a_i}$

Now, computing each of the $\Delta_{a_i}$ for i=1, 2…5 for given *U/D = {{o1, o2}, {o3, o4, o5, o6}}*

$\Delta_{a_1} = \frac{1}{|U/D|}\sum_{X \in U/D} (w1 \frac{|\underline{P}X|}{|U|} + w2\frac{|BN_R(X)|}{|U|})$

$\Delta_{a_1} = \frac{1}{2}\left(0.3 * \frac{0}{6} + 0.7 * \frac{4}{6}\right) + \frac{1}{2}\left(0.3 * \frac{2}{6} + 0.7 * \frac{4}{6}\right) = 0.5166$

$\Delta_{a_2} = \frac{1}{2}\left(0.3 * \frac{2}{6} + 0.7 * \frac{2}{6}\right) + \frac{1}{2}\left(0.3 * \frac{3}{6} + 0.7 * \frac{2}{6}\right) = 0.3583$

$\Delta_{a_3} = \frac{1}{2}\left(0.3 * \frac{0}{6} + 0.7 * \frac{6}{6}\right) + \frac{1}{2}\left(0.3 * \frac{0}{6} + 0.7 * \frac{6}{6}\right) = 0.7000$

$\Delta_{a_4} = \frac{1}{2}\left(0.3 * \frac{0}{6} + 0.7 * \frac{3}{6}\right) + \frac{1}{2}\left(0.3 * \frac{3}{6} + 0.7 * \frac{3}{6}\right) = 0.4250$

$\Delta_{a_5} = \frac{1}{2}\left(0.3 * \frac{0}{6} + 0.7 * \frac{5}{6}\right) + \frac{1}{2}\left(0.3 * \frac{1}{6} + 0.7 * \frac{5}{6}\right) = 0.6083$

Complete span = $\Delta_{R,D}^{Complete} = \frac{1}{|R|}\sum_{a_i \in R} \Delta_{a_i,D}$

$\Delta_{R,D}^{Complete} = \frac{1}{|5|}(0.5166 + 0.3583 + 0.7000 + 0.4250 + 0.6083) = 0.5216$

Now, computing each of the $\Delta_{a_i}$ for i=1, 2…5 for given *U/D = {{o1, o3, o4}, {o2, o5, o6}}*

$\Delta_{a_1} = \frac{1}{|U/D|}\sum_{X \in U/D} (w1 \frac{|\underline{P}X|}{|U|} + w2\frac{|BN_R(X)|}{|U|})$

$\Delta_{a_1} = \frac{1}{2}\left(0.3 * \frac{2}{6} + 0.7 * \frac{4}{6}\right) + \frac{1}{2}\left(0.3 * \frac{0}{6} + 0.7 * \frac{4}{6}\right) = 0.5166$

$\Delta_{a_2} = \frac{1}{2}\left(0.3 * \frac{2}{6} + 0.7 * \frac{3}{6}\right) + \frac{1}{2}\left(0.3 * \frac{1}{6} + 0.7 * \frac{3}{6}\right) = 0.4250$

$\Delta_{a_3} = \frac{1}{2}\left(0.3 * \frac{0}{6} + 0.7 * \frac{6}{6}\right) + \frac{1}{2}\left(0.3 * \frac{0}{6} + 0.7 * \frac{6}{6}\right) = 0.7000$

$\Delta_{a_4} = \frac{1}{2}\left(0.3 * \frac{0}{6} + 0.7 * \frac{6}{6}\right) + \frac{1}{2}\left(0.3 * \frac{0}{6} + 0.7 * \frac{6}{6}\right) = 0.7000$

$\Delta_{a_5} = \frac{1}{2}\left(0.3 * \frac{0}{6} + 0.7 * \frac{6}{6}\right) + \frac{1}{2}\left(0.3 * \frac{0}{6} + 0.7 * \frac{5}{6}\right) = 0.6417$

Complete span = $\Delta_R^{Complete} = \frac{1}{|R|} \sum_{a_i \in R} \Delta_{a_i}$

$\Delta_{R,D}^{Complete} = \frac{1}{|5|} (0.5166 + 0.4250 + 0.7000 + 0.7000 + 0.6417) = 0.7458$

From these computations it is seen that a decision D taken by experts leads to a higher span than the decision D1 by the experts. This can be understood as the fact that a decision has to be made by experts using the concept of span which is the covering ability of a subset. Here, both decision classes are considered. Span of this decision system can be considered as the ability of spanning ability of decision, that it includes as many of essential concepts and covers as much as possible in the "possibly covered" concepts.

If, however, the aim was to choose only one team, which has to be send to rescue mission, the span from Definition 1 (Yadav et al., 2019) suits well. The computation for span to determine best rescue team for flood relief by Definition 1 is as follow.

Let X = {o1, o2}

$\delta_{a_1} = \left(0.3 * \frac{0}{6} + 0.7 * \frac{4}{6}\right) = 0.4666$

$\delta_{a_2} = \left(0.3 * \frac{2}{6} + 0.7 * \frac{2}{6}\right) = 0.3333$

$\delta_{a_3} = \left(0.3 * \frac{0}{6} + 0.7 * \frac{6}{6}\right) = 0.7000$

$\delta_{a_4} = \left(0.3 * \frac{0}{6} + 0.7 * \frac{3}{6}\right) = 0.3500$

$\delta_{a_5} = \left(0.3 * \frac{0}{6} + 0.7 * \frac{5}{6}\right) = 0.5833$

$\delta_{R,X}^{Complete} = \frac{1}{|R|} \sum_{a_i \in R} \delta_{a_i} = \frac{1}{5}(0.4666 + 0.3333 + 0.7 + 0.35 + 0.58) = 0.4859$

And for Y = {o1, o3, o4}

$\delta_{a_1} = \left(0.3 * \frac{2}{6} + 0.7 * \frac{4}{6}\right) = 0.5666$

$\delta_{a_2} = \left(0.3 * \frac{2}{6} + 0.7 * \frac{3}{6}\right) = 0.4500$

$\delta_{a_3} = \left(0.3 * \frac{0}{6} + 0.7 * \frac{6}{6}\right) = 0.7000$

$\delta_{a_4} = \left(0.3 * \frac{0}{6} + 0.7 * \frac{6}{6}\right) = 0.7000$

$\delta_{a_5} = \left(0.3 * \frac{0}{6} + 0.7 * \frac{6}{6}\right) = 0.7000$

$\delta_{R,Y}^{Complete} = \frac{1}{|R|} \sum_{a_i \in R} \delta_{a_i} = \frac{1}{5}(0.5666 + 0.45 + 0.7 + 0.7 + 0.7) = 0.6233$

These computations using span of an information system leads to similar conclusion as of span of a decision system, wherein the second decision is given a better scoring than the first decision. Hence, is selected as expert, if it is a question of choice between these two sets. As can be seen these consider the properties of 'no food', 'no water', 'no utility' to mention a few as against first choice which considers much lesser coverage than the second solution.

Further, weights can be assigned to decision classes and weighted span of a decision system is defined as follows.

**Definition 6.** Given a decision system (*U, R, D*), $P \subseteq R$, $U/D = \{D1, D2, \ldots, Dr\}$. Define the weighted span of D as the weighted average,

$$\Delta_{P,D,u} = \sum_{X \in U/D} ui \left( w1 \frac{|\underline{P}X|}{|U|} + w2 \frac{|BN_R(X)|}{|U|} \right), w1, w2 \in [0,1], w1 + w2 = 1, \sum_i ui = 1$$

The main advantage of this definition of span is that in real life problems certain decisions categories are biased, means more important than other. For example, in the above example of sending relief to flood affected areas, the two teams can be urgent team, which is ready to be send quickly, while other team is the team which may take 24 hours to get ready. Hence weights can be assigned to each member of the indiscernibility relation of the decision class.

The weighted span of a decision system defines the ability of the decision attributes to cover the knowledge represented by the knowledge base of the decision system, given the importance level of each category in the decision table. The complete span of a weighted decision system is proposed as follows.

**Definition 7.** The complete span of a weighted decision system (*U, R, D*) based on the given attribute set is given by:

$$\Delta_{R,D,w}^{Complete} = \frac{1}{|R|} \sum_{a_i \in R} \Delta_{a_i,D,w}$$

Computations similar to as above can be performed for weighted decision classes.

Finding all subsets of the universal set is a NP complexity problem. And hence the solution can be found by approximate solutions as proposed in Yadav et al (2019). Making a decision system and varying various parameters involved with help of computing devise and can help find solutions in fraction of time, as against brute force decision making, which cannot on one go analyse all the data properties.

# 5. Some Mathematical Proofs on Span of Decision System

This section discusses some mathematical property of span of a decision system for two reducts of an information system.

**Proposition 1.** Suppose (U, R, D) be a decision system. Let attribute subsets R1, R2 $\subseteq$ R, be two reducts of the given decision system, then the span of R1 and span of R2 is same, i.e. $\Delta_{R2,D} = \Delta_{R1,D}$. Hence, for a span of (U, R, D), computed by two different reducts is same in a given decision system.
**Proof:** Since R1 and R2 are two reducts of the given decision system. Hence,

$POS_{R1}(D) = POS_{R2}(D)$, where $POS_P(D) = \cup \{\underline{P}X : X \in U/D\}$.
Hence, from definition it follows that,

$\cup \{\underline{R2}X : X \in U/D\} = \cup \{\underline{R1}X : X \in U/D\}$
Let $U/D = \{X1,\ldots,Xa\}$, $U/R1 = \{Y1,\ldots Yb\}$, $U/R2 = \{Z1,\ldots Zc\}$
Then by definition of equivalence classes, $Xi \cap Xj = \varphi$ and using the fact that
$\underline{R1}Xi \subseteq Xi$, $\underline{R1}Xj \subseteq Xj$. It follows that $\underline{R1}Xi \cap \underline{R1}Xj = \varphi$
Hence, $|\cup \{\underline{R1}X : X \in U/D\}| = \sum_{X \in U/D} \underline{R1}X$
Also, $\cup \{\underline{R2}X : X \in U/D\} = \cup \{\underline{R1}X : X \in U/D\}$ and $\cup \{X : X \in U/D\} = U$
Therefore,

$\Rightarrow \sum_{X \in U/D} \underline{R2}X = \sum_{X \in U/D} \underline{R1}X$ .....(1)

$\Rightarrow \frac{1}{|U/D|}\sum_{X \in U/D} (w1 \frac{|\underline{R1X}|}{|U|}) = \frac{1}{|U/D|}\sum_{X \in U/D} (w1 \frac{|\underline{R1X}|}{|U|})$ .....(2)

Also,
$\cup \{\overline{R1}X : X \in U/D\} = U$, $\cup \{\overline{R2}X : X \in U/D\} = U$, .......(3)
$|\cup \{\overline{R1}X : X \in U/D\}| = |\cup \{\overline{R2}X : X \in U/D\}|$
Also,
$\cup \{\overline{R1}X : X \in U/D\} = \cup \{\underline{R1}X : X \in U/D\} \cup \{Yj : Yj \notin \cup \{\underline{R1}X : X \in U/D\}\}$
$\cup \{\overline{R2}X : X \in U/D\} = \cup \{\underline{R2}X : X \in U/D\} \cup \{Zj : Zj \notin \cup \{\underline{R2}X : X \in U/D\}\}$

Using (3) we get,
$|\cup \{\underline{R1}X : X \in U/D\} \cup \{Yj : Yj \notin \cup \{\underline{R1}X : X \in U/D\}\}| = |U|$
$= |\cup \{\underline{R2}X : X \in U/D\} \cup \{Zj : Zj \notin \cup \{\underline{R2}X : X \in U/D\}\}| = |U|$

$\Rightarrow |\cup \{Yj : Yj \notin \cup \{\underline{R1}X : X \in U/D\}\}| = |\cup \{Zj : Zj \notin \cup \{\underline{R2}X : X \in U/D\}\}|$
(using (2) and the fact the union is disjoint) .....(4)

$\sum_{X \in U/D} \overline{R1}X = \sum_{X \in U/D} \underline{R1}X + |\cup BN_{R1}(X)|$,

$\sum_{X \in U/D} \overline{R2}X = \sum_{X \in U/D} \underline{R2}X + |\cup BN_{R2}(X)|$,

Using (6.4) and the fact that $BN_{R1}(X) = \cup \{Yj : Yj \notin \cup \{\underline{R1}X : X \in U/D\}\}$
$BN_{R2}(X) = \cup \{Zj : Zj \notin \cup \{\underline{R2}X : X \in U/D\}\}$.
Further, each of these Yj's and Zj's are disjoint.

$\Rightarrow \sum_{X \in U/D} |BN_{R1}(X)| = \sum_{X \in \frac{U}{D}} |BN_{R2}(X)|$
. ..........(5)

From (1) and (5) the following holds:
For a decision system,
$\Delta_{R1,D} = \frac{1}{|U/D|}\sum_{X \in U/D} (w1 \frac{|\underline{R1X}|}{|U|} + w2\frac{|BN_{R1}(X)|}{|U|})$.
$= \frac{1}{|U/D|}\sum_{X \in U/D} (w1 \frac{|\underline{R1X}|}{|U|}) + \frac{1}{|U/D|}\sum_{X \in U/D} (w2 \frac{|BN_{R1}(X)|}{|U|})$

And,
$\Delta_{R2,D} = \frac{1}{|U/D|}\sum_{X \in U/D} (w1 \frac{|\underline{R2X}|}{|U|} + w2\frac{|BN_{R2}(X)|}{|U|})$
$= \frac{1}{|U/D|}\sum_{X \in U/D} (w1 \frac{|\underline{R2X}|}{|U|}) + \frac{1}{|U/D|}\sum_{X \in U/D} (w2 \frac{|BN_R2(X)|}{|U|})$

$$\Delta_{R2,D} = \Delta_{R1,D}$$

Hence proved.

# 6. Conclusion and Future Works

This paper proposes a theoretical concept of span of a decision table. The application are illustrated with worked out examples and the novel concept is compared with the base paper definition. New definitions are provided for span and various real-life applications span of information tables and decision tables are provided in this paper. The paper also provides the proof of property that span of a decision table is invariant under the change of the features set among reducts. In future this work is can be extended on real time flood relief data, where this decision-making module can be provided as a software to experts. This indeed helps experts in taking right decision in constrained time limits.